\documentclass[letterpaper, 10 pt, conference]{ieeeconf} 
\IEEEoverridecommandlockouts                            
\usepackage{amsmath} 
\usepackage{amssymb} 
\usepackage{siunitx}

\usepackage[labelformat=simple,font=footnotesize,labelsep = period]{subcaption}
\usepackage{graphicx}
\usepackage{booktabs}
\usepackage{multirow}
\usepackage{stfloats}
\usepackage{cuted}
\usepackage{hyperref}

\newcommand{\tb}[3]{\setlength{\tabcolsep}{#2mm}\begin{tabular}{#1}#3\end{tabular}}

\usepackage{xcolor}

\title{\LARGE \bf
BatVision: Learning to See 3D Spatial Layout with Two Ears}

\author{\authorblockN{Jesper Haahr Christensen} 
\authorblockA{%\textit{DTU Electrical Engineering} \\
{Technical University of Denmark}\\
{\tt\small jehchr@elektro.dtu.dk}}
\and
\authorblockN{Sascha Hornauer} 
\authorblockA{
{UC Berkeley / ICSI}\\
{\tt\small saschaho@icsi.berkeley.edu}}
\and
\authorblockN{Stella X. Yu} 
\authorblockA{
{UC Berkeley / ICSI}\\
{\tt\small stellayu@berkeley.edu} }
}

\begin{document}

% For arxiv file
\thispagestyle{empty}
\newpage
\onecolumn
\begin{center}
This paper has been accepted for publication in 2020 International Conference on Robotics and Automation (ICRA).
\vspace{0.75cm}\\
DOI: \\ % \href{https://doi.org/10.1109/ICRA.2019.8793718}{\textcolor{blue}{10.1109/ICRA.2019.8793718}}\\
IEEE Xplore: \\% \href{https://ieeexplore.ieee.org/document/8793718}{\textcolor{blue}{https://ieeexplore.ieee.org/document/8793718}}
\vspace{1.25cm}
\end{center}
©2020 IEEE. Personal use of this material is permitted. Permission from IEEE must be obtained for all other uses, in any current or future media, including reprinting/republishing this material for advertising or promotional purposes, creating new collective works, for resale or redistribution to servers or lists, or reuse of any copyrighted component of this work in other works.
\twocolumn

\thispagestyle{empty}
\pagestyle{empty}

\maketitle

\begin{figure*}[b!]
\includegraphics[width=0.99\textwidth]{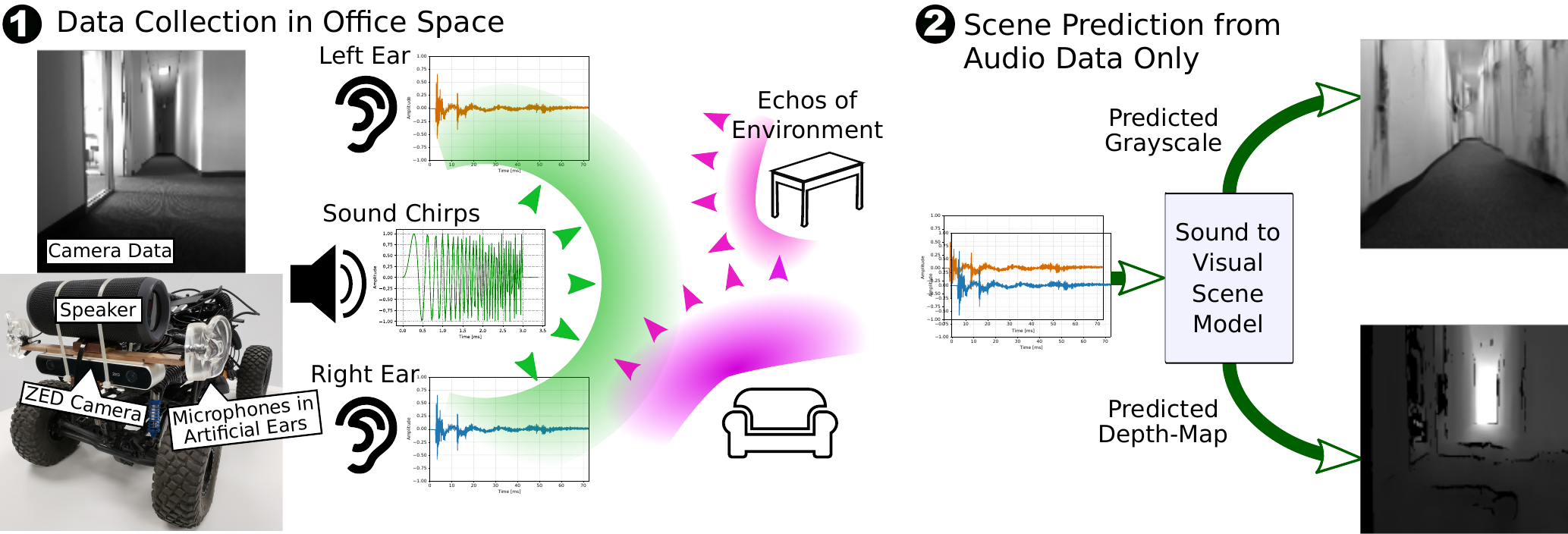}
\caption{Our {\it Batvision} system learns to generate visual scenes by just listening to echos with two ears.  Mounted on a model car, our system has two microphones embedded into artificial human ears, a speaker, and a stereo camera which is {\it only used during training} for providing visual image ground-truth.  {\bf 1)} The speaker emits sound chirps in an office space and the microphones receive echos returned from the environment.  The camera captures stereo image pairs, based on which depth maps can be calculated.  {\bf 2)} We train a model to turn binaural signals into visual scenes such as depth-maps or grayscale images.  Our results show surprisingly accurate reconstruction of the 3D spatial layout of indoor scenes from the input sound alone.
\label{fig:overview}}
\end{figure*}

\begin{abstract}
Many species have evolved advanced non-visual perception while artificial systems fall behind.  Radar and ultrasound complement camera-based vision but they are often too costly and complex to set up for very limited information gain.  In nature, sound is used effectively by bats, dolphins, whales, and humans for navigation and communication.  However, it is unclear how to best harness sound for machine perception.

Inspired by bats' echolocation mechanism, we design a low-cost {\it BatVision} system that is capable of seeing the 3D spatial layout of space ahead by just listening with two ears.  Our system emits short chirps from a speaker and records returning echoes through microphones in an artificial human pinnae pair.  During training, we additionally use a stereo camera to capture color images for calculating scene depths.  We train a model to predict depth maps and even grayscale images from the sound alone.  During testing, our trained BatVision provides surprisingly good predictions of 2D visual scenes from two 1D audio signals.  Such a sound to vision system would benefit robot navigation and machine vision, especially in low-light or no-light conditions.  Our code and data are publicly available.
\end{abstract}
%\begin{keywords}
%binaural echolocation, sound to vision, depth reconstruction
%\end{keywords}

\section{Introduction}
\label{sec:intro}
Our task is to train a machine learning system that can turn binaural sound signals to visual scenes.  Solving this challenge would benefit robot navigation and machine vision, especially in low-light or no-light conditions.  

While many animals sense the spatial layout of the world through vision, some species such as bats, dolphins, and whales rely heavily on acoustic information.  For example, bats have advanced ears that give them a form of vision in the dark known as {\it echolocation}: They sense the world by continuously emitting ultrasonic pulses and processing echos returned from the environment.  

It is indeed possible to locate highly reflecting ultrasonic targets in the 3D space by using an artificial pinnae pair of bats, which acts as complex direction dependent spectral filters {\it and} using head related transfer functions ~\cite{Biomimetic_Sonar,3DEcholocation}.

Likewise, humans who suffer from vision loss have shown to develop capabilities of echolocation using palatal clicks similar to dolphins, learning to sense obstacles in the 3D space by listening to the returning echoes ~\cite{human_echo,HumanUltrasonicEcholocation}.

Inspired by bats' echolocation, we design {\it BatVision} that can form a visual image of the 3D world by just listening to the environmental echo sound with two ears (Fig. \ref{fig:overview}).  

Contrary to existing works ~\cite{Biomimetic_Sonar,3DEcholocation}, our system uses only \textit{two} simple low-cost consumer-grade microphones to keep it small, mobile, and easily reproducible.
Our microphones are embedded into a human pinnae pair to utilize the spectral filters of an emulated human auditory system, which has an additional benefit of easy debugging by human engineers.

Mounted on a model car, our BatVision also has a speaker and a camera which is {\it only used during training} for providing visual image ground-truth.  Like bats, our speaker emits frequency modulated chirps in the audible spectrum, and our microphones receive echos returned from the environment.  Our camera captures stereo image pairs of the scene ahead, from which depth disparity maps can be calculated. 

During training, we first collect a dataset of time-synchronized binaural audio signals and stereo image pairs in an indoor office environment, and  then train a neural network model to predict images such as depth maps and grayscale images from audio data alone.  

During testing, we just need the sound signals to reconstruct depth maps or grayscale images.  By just listening with two ears, which receive sound echos at only two points in the 3D space, our BatVision is able to generate a depth map of the 3D space ahead that resolves features such as walls, hallways, door openings, and roughly outlined furniture correctly in azimuth, elevation, and distance, whereas our reconstructed grayscale images show surprisingly plausible floor layouts even though obstacles lack finer details. 

For a navigation system, such an intelligent sound system could provide information complementary to vision sensors, independent of light and at very low additional costs.  Our approach is conceptually simple, practically easy to implement,  and readily deployable on embedded mobile platforms. 

To the best of our knowledge, our BatVision is the first work that generates scene depth maps from binaural sound only.  Our code, model, and data are available at \url{https://github.com/SaschaHornauer/Batvision}.

%\begin{enumerate}
%\item Neural network architectures to learn a mapping from audio signals to virtual images. We show successful prediction of disparity-depth maps and plausible grayscale "scene-ahead" layout (free space or occlusions).
%\item A comparison of two input encodings, commonly used to map an audio signal into a rich latent feature space: raw waveforms and amplitude spectrograms. 
%\item A recorded dataset with about \SI{52000}{} audio chirps and returning echoes, captured in an indoor office space, synchronized and matched in time with RGB-D data.
%\end{enumerate}
%In all input/output configurations we compare generating predictions with and without an adversarial discriminator~\cite{Goodfellow:2014:GAN:2969033.2969125}. 

\section{Related Works}
\label{sec:review}

\noindent
\textbf{Biosonar Imaging and Echolocation.} 
Inspired by echolocation in animals, several papers \cite{HumanUltrasonicEcholocation,3DEcholocation, Bat-like_robot, BatSLAM, 3DbiosonarSceneAnalysis}  study target echolocation in the 2D or 3D space using ultrasonic frequency modulated (FM) chirps between $\SI{20}-\SI{200}{\kilo\hertz}$.  Bats emit pulse trains of very short durations (typically $<\SI{5}{\milli\second}$) and use received echoes to perceive their surroundings. 

In \cite{BatSLAM,HumanUltrasonicEcholocation},
microphones are placed in an artificial bat pinnae to receive the sound signal.  The natural form of the bat pinnae  acts as a frequency filter, useful for separating spatial information in both azimuth and elevation~\cite{bat3D,3DEcholocation}.  These works motivate our use of short FM chirps and artificial human pinnaes with integrated microphones.

In \cite{BatSLAM}, the task is to recognize scenes from echo-cochleogram fingerprints and to create a topological map of the surrounding.  In \cite{Bat-like_robot}, the goal is to autonomously drive a mobile robot while mapping and avoiding obstacles using azimuth and range information from ultrasonic sensors.  They classify echo spectrograms into obstacles or not, biological objects or not, along a single scan line and without visual reconstruction of the scene.

In \cite{HumanUltrasonicEcholocation}, ultrasonic echoes are recorded, dilated, and played back to a human subject in the audible spectrum.  After initial training, human subjects were able to pick up echolocation abilities to estimate azimuth,  distance, and to some extent, elevation of targets. In \cite{3DbiosonarSceneAnalysis,3DEcholocation}, 3D targets are localized based on an array of microphones instead of binaural microphones.
\\
%Common for all above mentioned work, is first the active sound sensing and algorithmic approach for 2D or 3D echolocation by delayed time of arrival and/or matched filtering. %Furthermore, as echolocation suggests, this only seeks to map out location of single or multiple targets and does therefore not investigate the possible correspondence between the perceived sound information with the visual surroundings. 

\noindent
\textbf{Sound Source Localization.} 
In \cite{Audio-Visual, SoundSource,Audio-Visual-Distillation, Ephrat:2018:LLC:3197517.3201357, multisensory2018},  deep neural network models are trained to localize the source of the sound (e.g. a piano) in images or videos.   Remarkable results are obtained in a self-supervised learning framework, demonstrating the potential of learning associations between paired audio-visual data. 

In \cite{Audio-Visual-Distillation},  sound is localized using an acoustic camera \cite{AcousticCamera}, a hybrid audio-visual sensor that provides RGB video overlaid with acoustic sound, aligned in time and space.  All the works on sound localization receive sound signals passively.

 In \cite{EnhancedBinaural}, sound is localized using emulated binaural hearing, with a model of human ears and head related transfer functions.  They test azimuth from $\SI{0}{\degree}$ to $\SI{360}{\degree}$ at \SI{5}{\degree} resolution and test elevation from $\SI{-40}{\degree}$ to $\SI{90}{\degree}$ at \SI{10}{\degree} resolution.

In \cite{speech2gesture}, an audio monologue of a speaker is turned into visual gestures of the speaker's arms and hands, by translating audio clips into $2D$ trajectories of joint coordinates.  Our sound to vision decoder model is inspired by their cross-domain translation success.
\\
%Typically audio signals can both be processed as a time-domain signal with e.g. SoundNet~\cite{aytar2016soundnet} or it may be visually represented as amplitude spectrograms and processed using CNNs. To model temporal dependencies in \textit{e.g.} video frames with audio over time, LSTM cells are commonly used.
%\\
%\\
%\noindent
%\textbf{Underwater Sonar Imaging} \quad
%Underwater imaging carried out from surface vessels, remotely operated vehicles (ROVs) and autonomous underwater vehicles (AUVs) is typically done using sonar technology due to the lack of visability in water. Two types of systems are usually employed; side-scan sonars and multibeam echosounders. Both of these emit ultrasonic sound waves through water and record received echoes. The information obtained is most commonly in 2D in the form of direction of arrival (DoA) and time of flight (ToF), i.e. range and azimuth. Resolution of such sensors are enhanced by employing large linear hydrophone arrays with beamforming~\cite{beamforming} and can for side-scan systems employ similar methods as found in SAR~\cite{SAR} imaging by synthetically extending the length of receiving array~\cite{SAS}. Imaging is performed by simply showing the intensity of the received echoes on their respective locations. 

\noindent
\textbf{Acoustic Imaging.} 
In non-Line-of-Sight imaging \cite{NLOS_imaging}, a microphone and a speaker array are used to emit and record FM sound waves.  The sound waves are chirps in the audible spectrum from $\SI{2}{\hertz}-\SI{20}{\kilo\hertz}$, emitted to propagate to a wall, a hidden object, and back to the microphone array.  They demonstrate successful object reconstruction at a resolution limited by the receiving microphone array.  In contrast, we capture the complete scene of the 3D space ahead with a small system mounted on a mobile device.

%The work carried out in this project distinguishes itselfusing only one binaural microphone pair and one speaker. 

\begin{figure}[b]
    \centering
    \includegraphics[width=\linewidth]{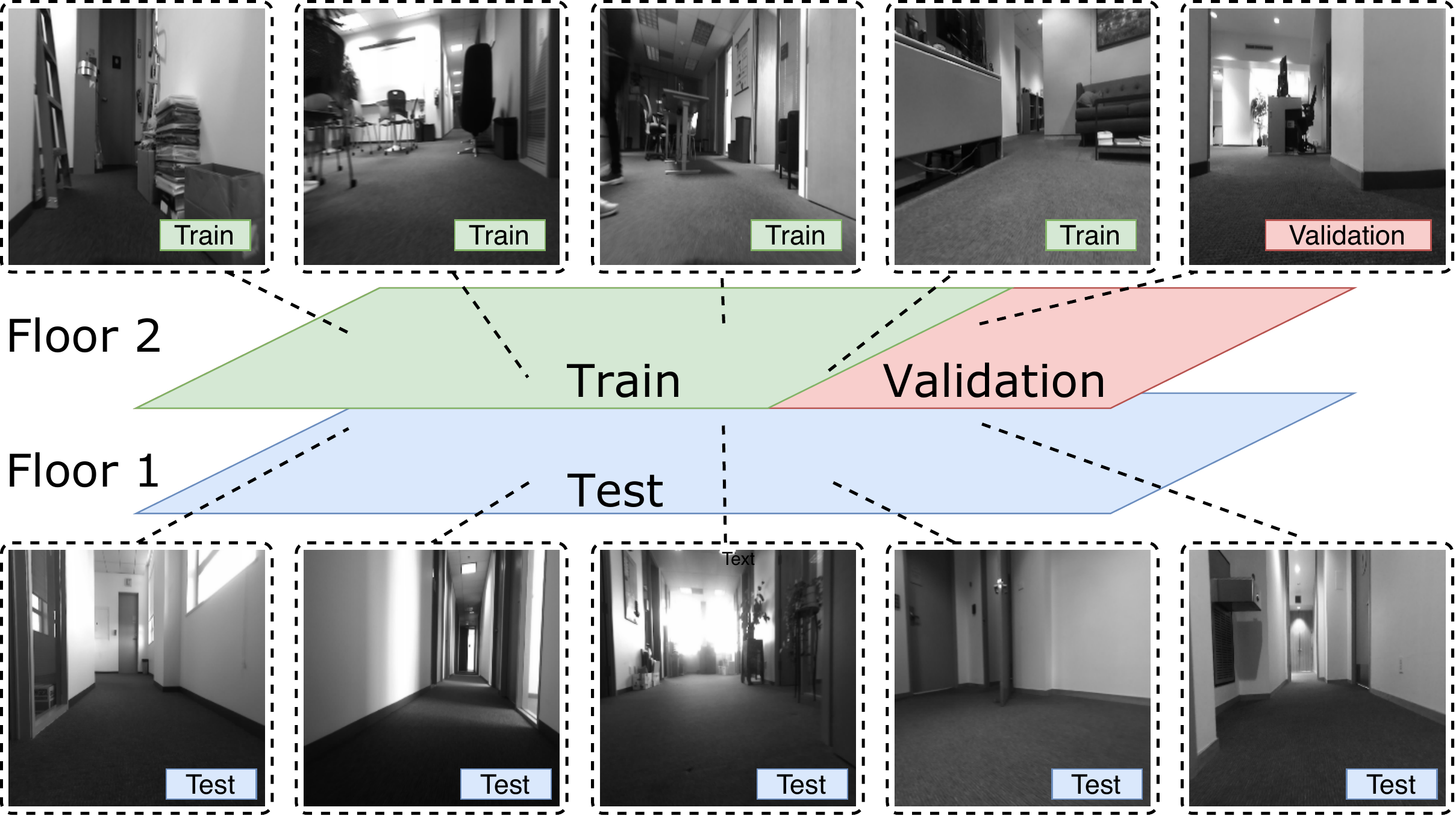}
    \caption{\textit{Data collected at different parts of the building are used for training, validation, and testing.}   Training and validation data are collected in separate areas of the same floor, whereas the test data come from another floor and have different obstacles and decorations.}
    \label{fig:dataset_comparison}
\end{figure}

\section{Collection of Our Audio-Visual Dataset} \label{sec:data}
We collect a new dataset of time-synchronized binaural audio, RGB images, and depth maps, which can be used for learning the associations between sound and vision.

\subsection{Our Data Collection Sites and Splits}
We traverse an office building in the hallways, open areas, conference rooms, and office spaces.   We fix our BatVision on a trolley and slowly push it around, so that there is no active motor noise corrupting our sound.  

We collect data at various spatial locations to  minimize correlation and maximize scene diversity (Fig~\ref{fig:dataset_comparison}).  A total of 39,500 and 7,500 instances at two different parts of the same floor are collected for training and validation respectively, with additional 5,040 instances on a different floor for testing.  While hallways appear similar, their spatial layout, furniture, occupancy, and decorations are different.

\subsection{Our Hardware: Speaker, Ears, and Camera}
We use a consumer-grade JBL Flip4 Bluetooth speaker to send out linear FM waveform chirps every half second (Fig. \ref{fig:overview}).  Each chirp sweeps from $\SI{20}{\hertz}-\SI{20}{\kilo\hertz}$ within a duration of \SI{3}{\milli\second}.
The waveform characteristics are designed %and combined into a continuous pulse train with \SI{500}{\milli\second} separation 
using the freely available software tool Audacity.
%\footnote{https://audacityteam.org/}

We adopt two low-cost consumer-grade omni-directional USB Lavalier MAONO AU-410 microphones, separated at 
 approximately \SI{23.5}{\centi\meter} apart. 
%These are capable of operating with a sampling frequency up to \SI{192}{\kilo\hertz} with \SI{24}{\textrm{bit}} resolution. 
Each microphone is mounted in a Soundlink silicone ear to effectively emulate an artificial human auditory system.  We record sound using \texttt{PyAudio}
%\footnote{http://people.csail.mit.edu/hubert/pyaudio/} 
for Python at \SI{44.1}{\kilo\hertz} and \SI{24}{} bits per sample.

We use a ZED  camera to capture stereo image pairs and extract depth maps from them.
Our camera, speaker, and artificial ears are mounted on a small model car (Fig.~\ref{fig:overview}). 

% \begin{figure}
%     \centering
%     \includegraphics[width=0.80\linewidth]{graphics/robot.JPG}
%     \caption{Photo of our mobile robot configured with artificial human pinnaes, microphones, speaker and a stereo camera. Note that at this stage the motor was not used to limit noise.}
%     \label{fig:robot}
% \end{figure}

% \begin{figure}
%     \centering
%     \includegraphics[width=\linewidth]{graphics/ICSI_floorplan.pdf}
%     \caption{Dataset train and validation split visualized on top of the ICSI floorplan.}
%     \label{fig:ICSI_floorplan}
% \end{figure}

% \begin{figure*}[t]
%     \begin{subfigure}[b]{4.5cm}
% 		\includegraphics[width=4.5cm,height=3cm,trim=0 12 0 12,left]{graphics/waveform.pdf}
%         \caption{}
%     \end{subfigure}
%     \hspace{-0.3cm}
%     \begin{subfigure}[b]{4.8cm}
%         \centering
% 		\includegraphics[width=4.8cm,height=3cm,trim=0 12 0 12]{graphics/spectrum.pdf}
%         \caption{}
%     \end{subfigure}
%     \hspace{-0.15cm}
%     \begin{subfigure}[b]{0.235\textwidth}
%         \centering
%         \includegraphics[width=\linewidth,height=3cm]{graphics/depth.PNG}
%         \caption{}
%     \end{subfigure}
%     \hspace{-0.01cm}
%     \begin{subfigure}[b]{0.235\textwidth}
%         \includegraphics[width=\linewidth,height=3cm,right]{graphics/20190814T195406_095922000_left.png}
%         \caption{}
%     \end{subfigure}
%     \caption{Data sample. (a) Recorded audio waveform for a left ear. (b) Spectrogram for left ear. (c) Depth map. Pixels with no range measurements are set 0. (d) RGB image from left camera.}
%     \label{fig:data_sample}
% \end{figure*}

\begin{figure}[!b]
\centering
\tb{c}{0}{
\includegraphics[width=\linewidth,trim=20 6 0 10]{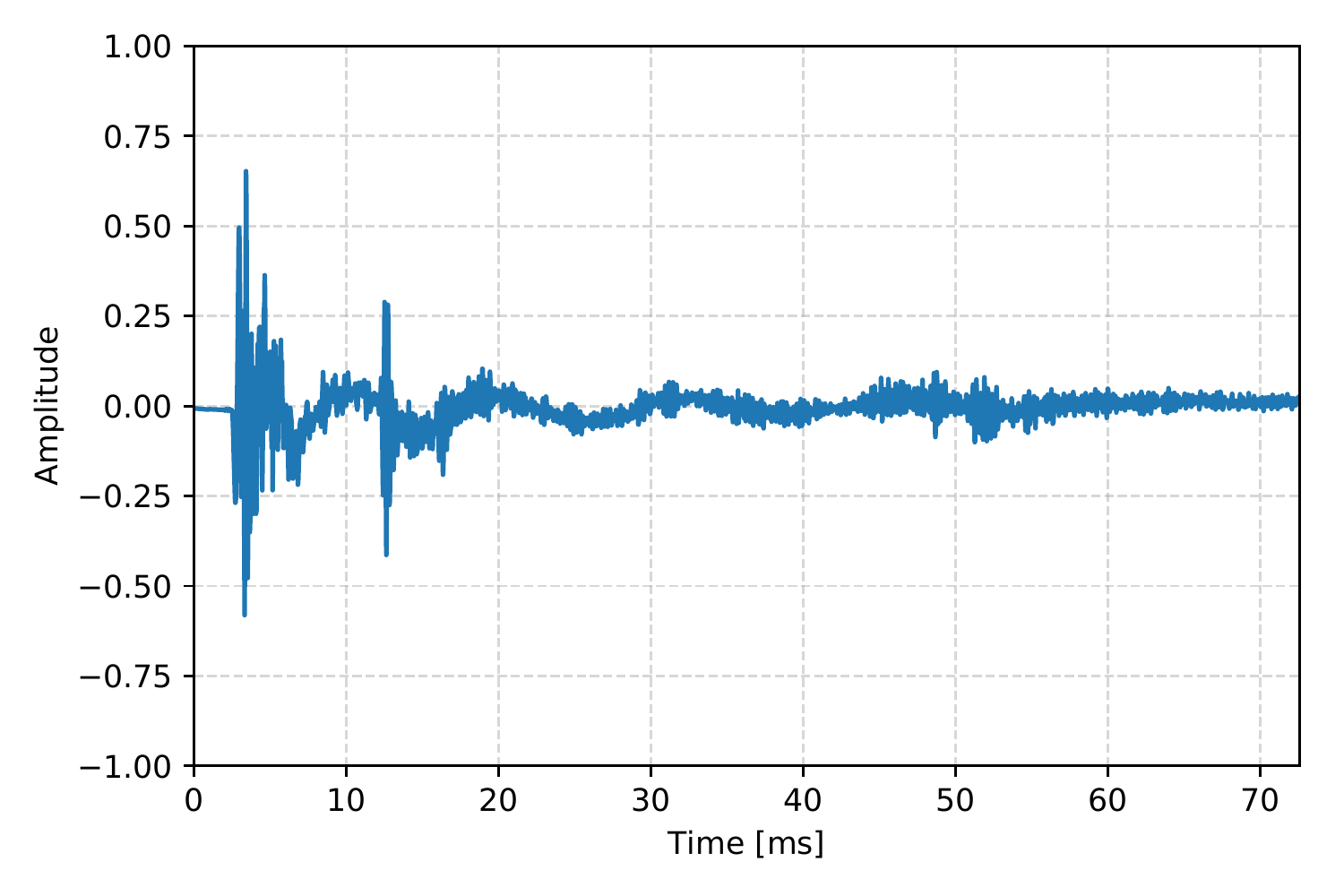}\\
\includegraphics[width=\linewidth,trim=20 6 0 10]{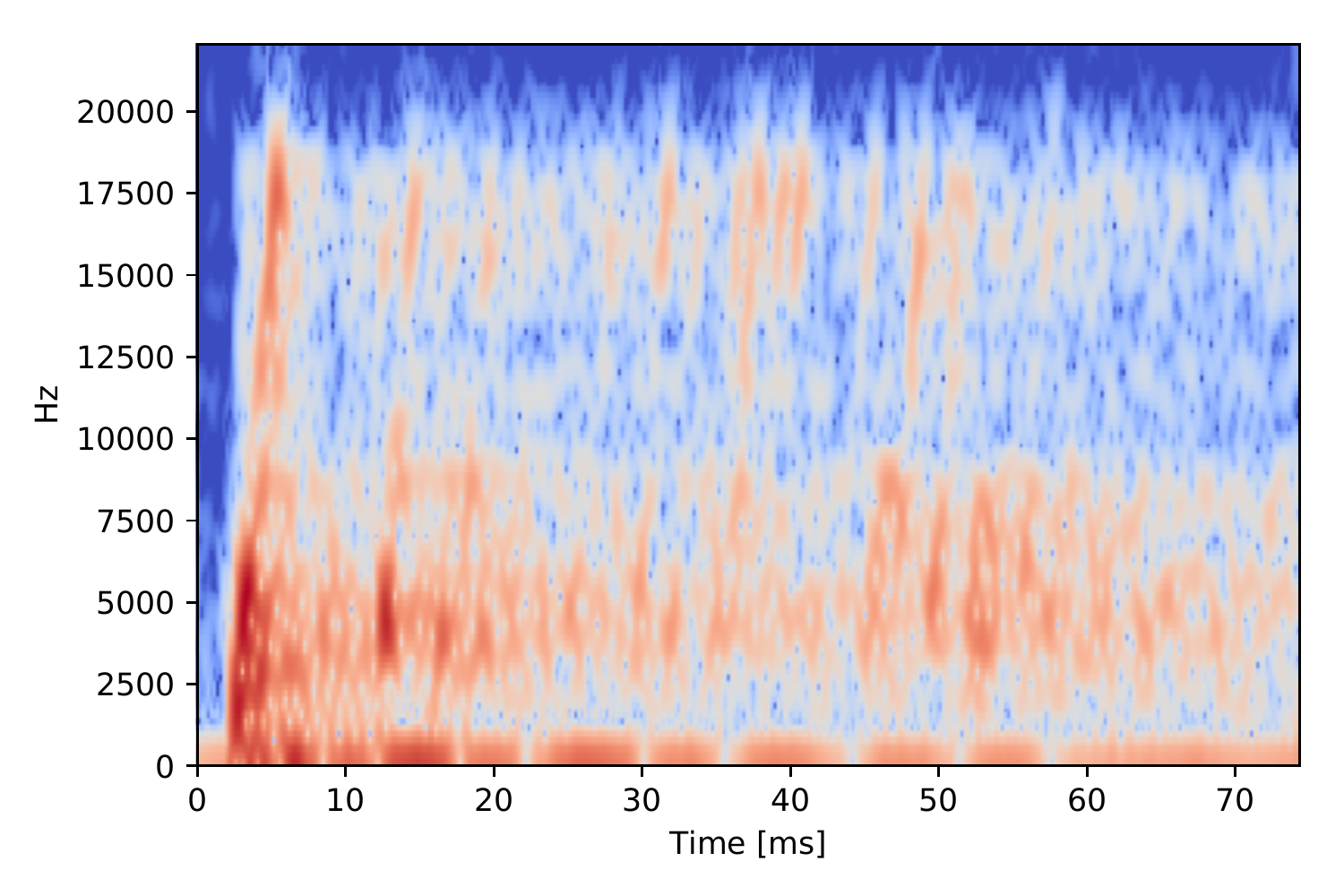}\\
}
\caption{\textit{Sample audio waveform and its amplitude spectrogram from a single microphone}.  The echo appears after the chirp (first peak) at about $\SI{3}{\milli\second}$.
    \label{fig:data_sample_audio}}
\end{figure}

\subsection{Our Audio clips and Visual Images}

% In order to strike a balance between receiving echos from objects at a relevant distance and reducing later echos from multipath-effects we choose the length of an audio sample to include echoes traveling up to \SI{12}{\meter}.
% To this end, we extract audio windows of \SI{3200}{} samples (corresponding to \SI{72.5}{\milli\second}) from the data. %These include the emitting chirp and received echoes. 

We choose the length of each audio instance to be \SI{72.5}{\milli\second}, so that it includes echoes traveling up to \SI{12}{\meter}.   This time window selection reflects a trade-off between receiving echos within the distance relevant for navigation and reducing later echos from multiple reflection paths.  Each of our audio clips has \SI{3200}{} frames, containing one chirp and returned echoes. 

We synchronize all the audio instances by the time of the recorded chirp.  However, during training,  we augment the audio data by jittering the position of the window by 30\%.

We consider two audio representations: $1D$ raw waveforms and $2D$ amplitude spectrograms.  The \texttt{LibROSA} library for Python is used to compute spectrograms with \SI{512}{} points for FFT and Hanning window size \SI{64}{}.  
Fig. ~\ref{fig:data_sample_audio}  shows the probing %$\SI{20}{\hertz}-\SI{20}{\kilo\hertz}$ 
chirp at $\SI{3}{\milli\second}$ and the returned echoes afterwards.  %Fig.~\ref{fig:data_sample_image} shows a RGB image and the depth map of that scene. Brighter intensities yield measurements of greater range. 

We compute the scene depth using the API of our camera, range clipped within \SI{12}{\meter}.  We normalize the depth value to be between 0 and 1.  Pixels where the camera is unable to produce a valid measurements are set to 0. 

% \begin{figure}
%     \centering
%     \begin{subfigure}{0.49\linewidth}
%         \centering
%         \includegraphics[width=\linewidth,height=3cm]{graphics/20190814T195406_095922000_left.png}
%         %\caption{}
%     \end{subfigure}
%     \begin{subfigure}{0.49\linewidth}
%         \centering
%         \includegraphics[width=\linewidth,height=3cm]{graphics/depth.PNG}
%         %\caption{}
%     \end{subfigure}
%     \caption{\textit{Data sample from the ZED camera.} \textbf{Left:} RGB image. \textbf{Right:} Computed depth map of the imaged scene. Brighter is further away. Pixels with no range measurements are set 0 (black).}
%     \label{fig:data_sample_image}
% \end{figure}

\section{Our Sound to Vision Prediction Models}\label{sec:audio2visual}

\begin{figure}[!h]
    \centering
    \includegraphics[width=1.0\linewidth]{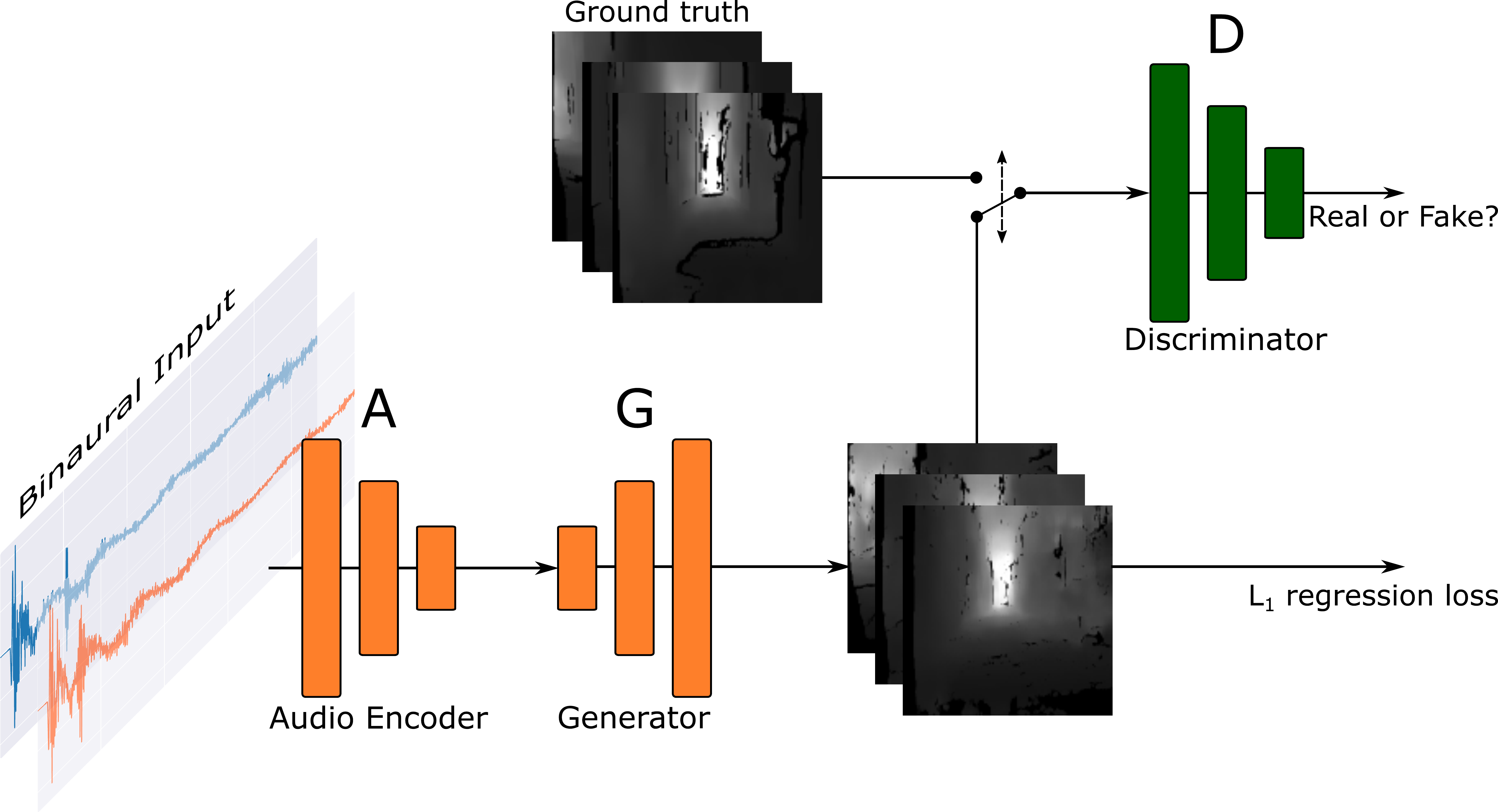}
    \caption{\textit{Our sound to vision network architecture.}  The temporal convolutional audio encoder $A$ turns the binaural input into a latent audio feature vector, based on which the visual generator $G$ predicts the scene depth map.  The discriminator $D$ compares the prediction with the ground-truth and enforces high-frequency structure reconstruction at the patch level.
    \label{fig:model}}
\end{figure}

We use an encoder-decoder network architecture to turn the audio clip into the visual image, and further improve the quality of generated images using an adversarial discriminator to contrast them against the ground-truth (Fig. \ref{fig:model}).  

We train our model with two possible audio representations.  Our experiments indicate that spectrograms yield slightly better sound-to-vision predictions over raw waveforms.  However, as we aim for a real-time BatVision system on embedded platforms, we focus on raw waveforms which are more computationally efficient.

\subsection{Our Audio Encoder $A$} \label{sec:audio_encoder}

%The output of the audio encoder enters the generator (Section~\ref{sec:generative_models}) which is to generate a $2D$ (or $3D$) image---no time, based on the 

\begin{figure}
    \centering
    \includegraphics[width=\linewidth]{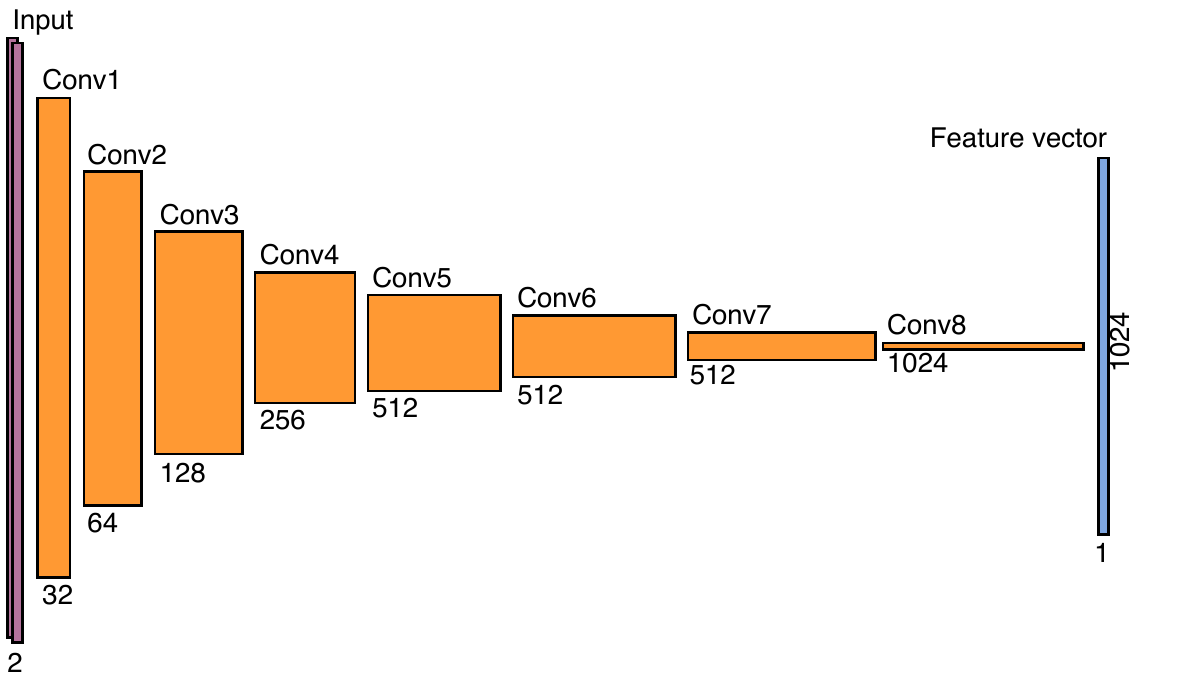}
    \caption{\textit{Our audio encoder for the raw waveform.}  We use 8 convolutional layers to turn the two-channel representations of the audio waveform into a 1024-dimensional feature vector.
    \label{fig:waveform_encoder}}
\end{figure}

%\\
\noindent
\textbf{Encoder for Waveforms.} 
%\subsubsection{Waveforms}
%\label{sec:waveforms}
Following  SoundNet~\cite{SoundNet}, we represent the binaural input as two channels of $1D$ signals and transform it into a 1024-dimensional feature vector with 8 temporal convolutions.  See Fig.~\ref{fig:waveform_encoder} and  Table~\ref{tab:audio_encoder} for details.

%As shown in Fig.~\ref{fig:waveform_encoder} we input the two $1D$ audio waveforms (binaural signal) and concatenate the input by the channel dimension (early fusion). Next follows a series of 8 temporal convolutions to downsample the signal into a final output of a 1024-dimensional feature vector. Details of our audio encoder are summarized in Table~\ref{tab:audio_encoder}. \\
%{\setlength{\belowcaptionskip}{-4ex}

\noindent
\textbf{Encoder for Spectrograms.} 
%\subsubsection{Spectrograms}
Likewise, with successive temporal convolutions and downsampling, we gradually reduce the time-dimension of the spectrograms down to 1, producing a $1 \times f \times 1024$ feature vector, where $f$ is the number of final frequencies.  $f$ depends on the downsampling factors along the y-axis of the spectrogram.

%Dependet on the downsampling factor of the frequency axis, the output dimension can take forms; as in Section~\ref{sec:waveforms} it can result in 1024 ($1\times 1 \times 1024$), or it can result in $f \times 1 \times 1024$ with $f$ being the frequency axis.

\begin{table}[]
\centering
\caption{Layer configuration of our waveform audio encoder}
\label{tab:audio_encoder}
\begin{tabular}{@{}lcccc@{}}
\toprule
Layer & \multicolumn{1}{l}{\# of Filters} & \multicolumn{1}{l}{Filter size} & \multicolumn{1}{l}{Stride} & \multicolumn{1}{l}{Padding} \\ \midrule
Conv1 & 32                                & 228                             & 2                          & 114                         \\
Conv2 & 64                                & 128                             & 3                          & 64                          \\
Conv3 & 128                               & 64                              & 3                          & 32                          \\
Conv4 & 256                               & 32                              & 3                          & 16                          \\
Conv5 & 256                               & 16                              & 3                          & 8                           \\
Conv6 & 512                               & 8                               & 3                          & 4                           \\
Conv7 & 512                               & 4                               & 3                          & 2                           \\
Conv8 & 1024                              & 3                               & 3                          & 1                           \\ \bottomrule
\end{tabular}
\end{table}

\subsection{Our Visual Image Generator $G$}
\label{sec:generative_models}
The generator decodes the latent audio feature vector and expands it into visual scene image.  For raw waveforms, successive deconvolutions yield the best results, whereas for spectrograms, a UNet-type encoder-decoder network~\cite{u-net} yields best results.  We investigate several resolutions for reconstructed images, from $16\times 16$ to $128\times 128$.\\%, shown in Fig. \ref{fig:no-gan}. \\

%\\
\noindent
\textbf{Decode by A UNet.} 
To transform the output of our audio encoder to a $2D$ image representation suitable for a UNet, we reshape the 1024-dimensional feature vector into a $32\times 32 \times 1$ tensor.  For spectrograms,  where the audio encoder outputs a $1\times f \times 1024$ vector and $f \neq 1$, we first apply two fully connected linear layers before reshaping it into a $32\times 32 \times 1$ tensor.  The output of this generator depends on the target resolution, \textit{e.g.} $128\times 128 \times 1$. 

The encoder of the UNet downsamples the $32\times 32 \times 1$ input through several layers of double convolutions followed by batch normalization and ReLU,
%convolution, batch normalization, ReLU, convolution, batch normalization and ReLU. 
whereas the decoder of the UNet upsamples the input through double de-convolutions followed by batch normalization and ReLU.
Skip connections are utilized wherever possible.%, \textit{i.e.} up to $32\times 32$ resolution.

\noindent
\textbf{Decode from Direct Upsampling.} 
%\subsubsection{Direct upsampling}
Given the $1\times 1 \times 1024$ latent audio vector,  we apply a series of upsampling layers (as in the  \textbf{UNet} decoder) to reach the target resolution. See the layer configuration for the $128\times 128 \times 1$ output  in Table~\ref{tab:direct_decoder}. 
%In case of the spectrogram encoder we employ the same strategy as in Section~\ref{sec:unet} to reach the correct input shape.

%The dimension of the bottleneck is $2\times 2 \times 512$ and the final output is depended on the target resolution, \textit{e.g.} $128\times 128 \times 1$. Our downsampling layers employ series of: convolution, batch normalization, leaky ReLU, convolution, batch normalization and leaky ReLU. The upsampling layers employ a similar series of with the first operation being a de-convolution (to up-sample) rather than a convolution. We utilize skip connections wherever possible, \textit{i.e.} up to $32\times 32$ resolution.

\begin{table}[]
\centering
\vspace{3mm}
\caption{Layer configuration of the direct upsampling generator for the $128\times 128$ image}
\label{tab:direct_decoder}
\begin{tabular}{@{}lccccc@{}}
\toprule
Layer & \multicolumn{1}{l}{\# of Filters} & \multicolumn{1}{l}{Filter size} & \multicolumn{1}{l}{Stride} & \multicolumn{1}{l}{Padding} & \multicolumn{1}{l}{Res.} \\ \midrule
Up1   & 512                               & 4                               & 1                          & 0                           & 4                        \\
Up2   & 512                               & 4                               & 2                          & 1                           & 8                        \\
Up3   & 256                               & 4                               & 2                          & 1                           & 16                       \\
Up4   & 128                               & 4                               & 2                          & 1                           & 32                       \\
Up5   & 128                               & 4                               & 2                          & 1                           & 64                       \\
Up6   & 64                                & 4                               & 2                          & 1                           & 128                      \\
Final & 1                                 & 1                               & 1                          & 0                           & 128                      \\ \bottomrule
\end{tabular}
\end{table}

\subsection{Our Adversarial Discriminator} \label{sec:adversarial}

We add an adversarial discriminator $D$ for generating more detailed and realistic predictions.
%conditioned on the difference between the output of the generator and the ground truth collected from the stereo camera. 
We implement the discriminator as a PatchGAN \cite{pix2pix} to ensure that the predicted visual image has similar looking patches as the set of ground-truth images;  $D$ tries to classify whether each $N\times N$ patch looks real or fake as a ground truth sample, where $N$ is roughly $1/3$ of the image size.
%We model our discriminator as series of 3 convolutions for depth map resolutions of $16 \times 16$, $32\times 32$ and $64\times 64$ pixels and 4 convolutions for a resolution of $128\times 128$ pixels. 
$D$ consists of a few convolutional layers with depth, kernel size and stride 
parameters dependent on the final output image size.  See the layer configuration in Table~\ref{tab:pathGAN} for size $128 \times 128$.

\begin{table}[]
\centering
\caption{PatchGAN discriminator configuration for $128\times 128$ images.}
\label{tab:pathGAN}
\begin{tabular}{@{}lcccc@{}}
\toprule
Layer & \multicolumn{1}{l}{\# of Filters} & \multicolumn{1}{l}{Filter size} & \multicolumn{1}{l}{Stride} & \multicolumn{1}{l}{Padding} \\ \midrule
Conv1 & 64                                & 4                               & 2                          & 1                           \\
Conv2 & 128                               & 4                               & 2                          & 1                           \\
Conv3 & 256                               & 4                               & 2                          & 1                           \\
Conv4 & 1                                 & 4                               & 2                          & 1                           \\ \bottomrule
\end{tabular}
\end{table}
\begin{figure*}
    \centering
    \vspace{3mm}
    \includegraphics[width=1\textwidth]{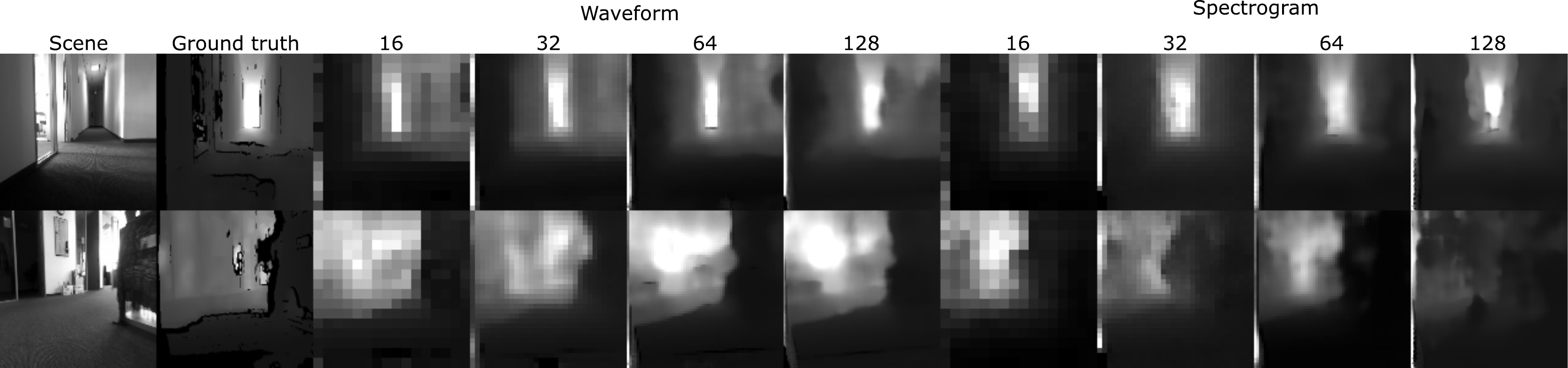}
    \caption{\textit{Sample sound-to-vision predictions by Generator $G$ only without the adversarial discriminator $D$}. Columns 1-2 show the grayscale scene image  and the ground-truth depth map.
    The rest columns show predictions from waveforms and spectrograms at size $16\times 16$, $32\times 32$, $64\times 64$, and $128\times 128$.
    \label{fig:no-gan}}
\end{figure*}

\section{Experimental Results}

\subsection{Generator Only Without Discriminator}
\label{sec:experiments_nogan}
In a preliminary study that compares input modes and fusion design choices, we predict small images at size $16\times 16$.  We have the following observations.
\begin{itemize}
\item For raw waveforms, early fusion (left-right-channel concatenation of the input audio) outperforms late fusion (concatenation at \texttt{Conv8}, see Fig.~\ref{fig:waveform_encoder}). 
\item Spectrograms yield slightly better results than raw waveforms. 
\end{itemize}
However, as we aim for real-time performance on embedded platforms,  we focus on the least computationally expensive method using waveforms. %However, we report obtained loss on both approaches for comparison.

%The audio encoder and generator learns to map the audio signals to a visual representation via a $L_1$ regression loss:
We compute the prediction error via an $L_1$ regression loss:
\begin{align}
    \mathcal{L}_{L_1}(G) = \mathbb{E}_{x,y}\left[||y-G(A(x))||_1\right]
\end{align}
where $x$ is the audio waveforms or spectrograms, $y$ is the ground truth visual image (depth map or grayscale scene image), $A$ is the audio encoder, and $G$ is the generator. 

We use leaky ReLU with slope $0.2$, batch size $16$, and Adam solver \cite{adam_optimizer} with an initial learning rate of $1\times 10^{-4}$ with parameters $\beta_1$ and $\beta_2$ set to $0.9$ and $0.999$ respectively. 

%We experimented with different depths, widths, kernel sizes and later fusion (\textit{e.g.} separate encoding tracks and concatenate in \texttt{Conv8}), combinations of both waveform and spectrogram features, down- and upsampling factors  and found the mentioned configuration in Section~\ref{sec:audio_encoder}--\ref{sec:generative_models} to give the best overall performance.

% In Table~\ref{tab:network_combinations} we report quantitative test results on a selected set of investigated audio encoder and generator configurations. The table further reports the results achieved by using the mean image of the training set and random inputs drawn from a uniform distribution on the interval $\left[0, 1\right)$. 

Table~\ref{tab:network_combinations} compares various model choices along with two trivial reconstruction baselines which do not learn any sound and vision associations at all:
\begin{enumerate}
\item The mean depth map of the training set.
\item Random uniform noise in the $\left[0, 1\right)$ range. 
\end{enumerate}
For raw waveforms, direct upsampling and early fusion perform the best. For spectrograms, early fusion, downsampling to $1\times 10 \times 1024$ and the UNet generator perform best.  
These two best configurations are retrained for output dimensions of $32\times 32$, $64\times 64$ and $128\times 128$, and the loss is higher for a larger depth map (Table~\ref{tab:results-gan}). 

Fig.~\ref{fig:no-gan} compares reconstructions at different resolutions.   Fig.~\ref{fig:image} shows more samples of diverse scenes at reconstruction size $128\times 128$.  The sound-to-vision predictions provide a rough outline of the spatial layout of the 3D scene.

\begin{table}[]
\centering
\caption{``Generator Only'' results for $16\times 16$ imgs. on the test set.}
\label{tab:network_combinations}
\begin{tabular}{@{}lcccc@{}}
\toprule
Audio Encoder                & Fusion                 & Shape                                     & Generator            & Loss            \\ \midrule
\multirow{4}{*}{Waveform}    & \multirow{2}{*}{Early} & \multirow{2}{*}{1024}                     & UNet                 & 0.0883          \\ \cmidrule(l){4-5} 
                             &                        &                                           & Direct               & \textbf{0.0838} \\ \cmidrule(l){2-5} 
                             & \multirow{2}{*}{Late}  & \multirow{2}{*}{1024}                     & UNet                 & 0.0894          \\ \cmidrule(l){4-5} 
                             &                        &                                           & Direct               & 0.0845          \\ \midrule
\multirow{4}{*}{Spectrogram} & \multirow{4}{*}{Early} & \multirow{2}{*}{1024}                     & UNet                 & 0.0834          \\ \cmidrule(l){4-5} 
                             &                        &                                           & Direct               & 0.0790          \\ \cmidrule(l){3-5} 
                             &                        & \multirow{2}{*}{$1\times 10 \times 1024$} & UNet                 & \textbf{0.0773} \\ \cmidrule(l){4-5} 
                             &                        &                                           & Direct               & 0.0778          \\ \midrule
Mean Depth                         & \multicolumn{1}{l}{}   & \multicolumn{1}{l}{}                      & \multicolumn{1}{l}{} & 0.1058          \\ \midrule
Random Noise                       & \multicolumn{1}{l}{}   & \multicolumn{1}{l}{}                      & \multicolumn{1}{l}{} & 0.3654          \\ \bottomrule
\end{tabular}
\end{table}

% \begin{table}[]
% \centering
% \caption{``No-GAN'' quantitative results for 32, 64, 128 waveform and spectrogram on the test set.}
% \label{tab:results_no-gan}
% \begin{tabular}{@{}lcccccc@{}}
% \toprule
% \multirow{2}{*}{Model} & \multicolumn{3}{c}{Waveform} & \multicolumn{3}{c}{Spectrogram} \\
%                       & 32       & 64      & 128     & 32       & 64        & 128      \\ \midrule
% Loss                   & 0.0852   & 0.0862  & 0.0880  & 0.722    & 0.0726    & 0.0742   \\ \bottomrule
% \end{tabular}
% \end{table}

\subsection{Generator with Adversarial Discriminator}
\label{sec:experiments_gan}

We use an Generative Adversarial network (GAN) model at the patch level to improve the visual reconstruction quality.
%The use of an adversarial discriminator improves depth-maps qualitatively with more accurate details even though the average loss increases slightly. Furthermore, based on audio, we reconstruct grayscale images which show an abstraction of how the room layout could look like with respect to free space and obstacles. Even though features of visual appearance are not present in the audio signal, we achieve plausible floor and wall layouts using the discriminator.
We use the following least-squares loss instead of a sigmoid cross-entropy loss in order to avoid vanishing gradients \cite{LSGAN}:  \begin{align}
&\hspace{-5pt}\mathcal{L}_{GAN}(D)\! =\! \mathbb{E}_{y}\left[\|1-D(y)\|^2_2\right]
\!+\!\mathbb{E}_{x}\left[\|D(G(A(x)))\|^2_2\right] \\
&\hspace{-5pt}\mathcal{L}_{GAN}(G)\!=\! \mathbb{E}_{x}\left[\|1-D(G(A(x)))\|^2_2\right]
\end{align}
%where $x$ is the left and right audio waveform, $y$ is the ground truth depth map from the stereo camera, $D$ is the discriminator, $G$ is the generator and $A$ is the audio encoder.
Our full objective is thus:
\begin{align}
    \min_G \max_D\quad \frac{1}{2} \mathcal{L}_{GAN}(D)+\mathcal{L}_{GAN}(G)+\lambda \mathcal{L}_{L_1}(G)
\end{align}
where $\lambda$ is a weight factor. 
We use leaky ReLU with slope 0.2, $\lambda = 100$, batch size $16$, and Adam solver with learning rate set to $2\times 10^{-4}$ with parameters $\beta_1$ and $\beta_2$ set to $0.5$ and $0.999$ respectively.

\begin{figure}[ht!]
    \centering
    \vspace{3mm}
    \includegraphics[width=\linewidth]{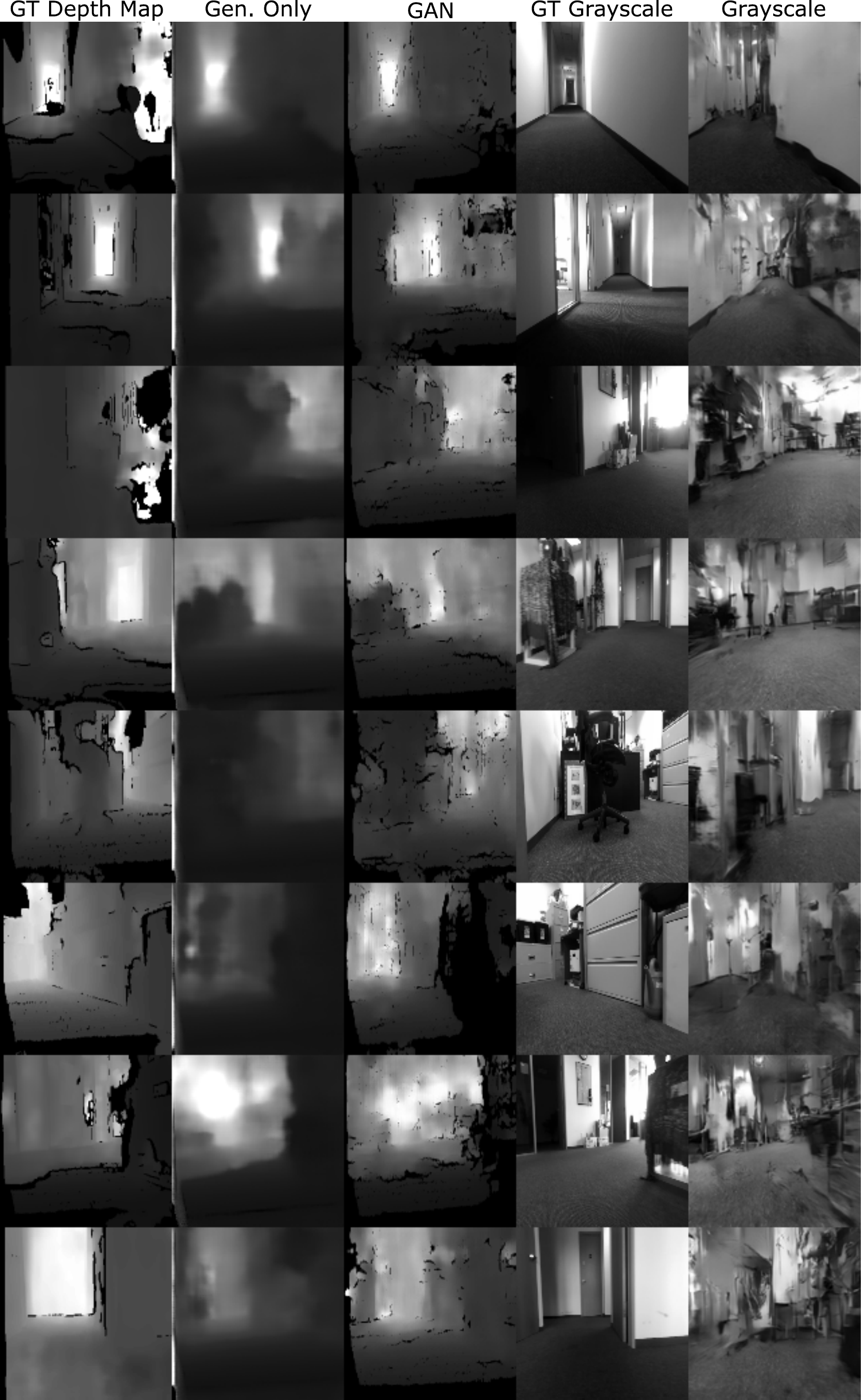}
    \caption{\textit{Good test sample reconstructions at the $128\times 128$ output resolution.}  Columns 1 and 4 show the ground truth depth map and grayscale scene image.  The remaining columns show predictions from  raw waveforms.  Overall, our generated depth maps show correct mapping of close and distant areas even for row 3, where errors are present in the ground-truth itself.
    \label{fig:image}}
\end{figure}

\begin{figure}[ht!]
    \centering
    \vspace{3mm}
    \includegraphics[width=\linewidth]{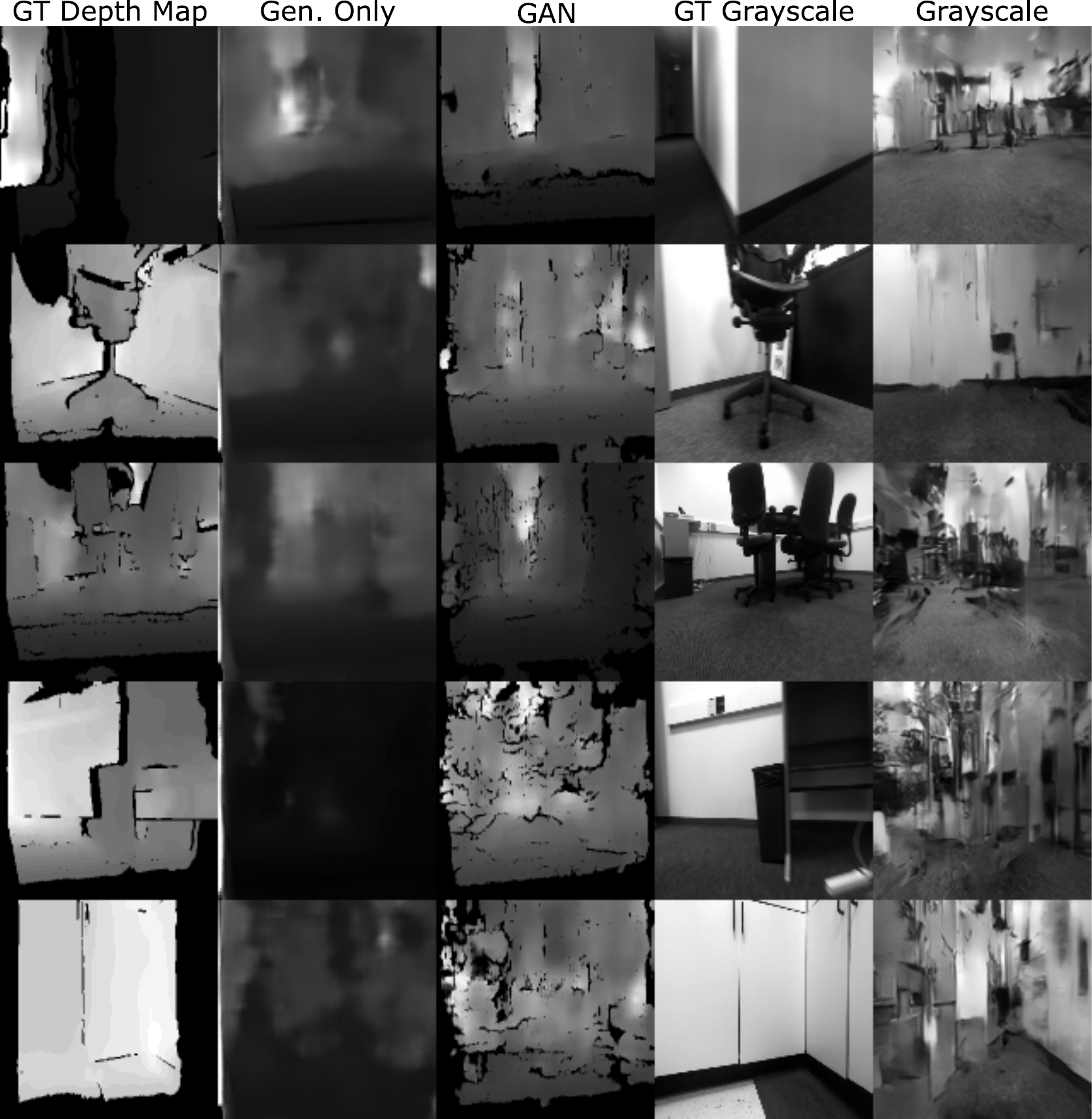}
    \caption{\textit{Poor test sample reconstructions}.  Same conventions as Fig. \ref{fig:image}. Up-close and complex objects are not well represented.}
    \label{fig:bad}
\end{figure}

\begin{table}[]
\centering
\vspace{3mm}
\caption{$L_1$ test loss for waveforms and spectrograms at resolution $32\times 32$, $64\times 64$, and $128\times 128$
}
\label{tab:results-gan}
\begin{tabular}{@{}lcccccc@{}}
\toprule
\multirow{2}{*}{Model} & \multicolumn{3}{c}{Waveform (D. Upsampling)}                                       & \multicolumn{3}{c}{Spectrogram (UNet Style)}                                    \\
                       & 32                   & 64                   & 128                  & 32                   & 64                   & 128                  \\ \midrule
\textit{Gen. Only}                 & \multicolumn{1}{l}{} & \multicolumn{1}{l}{} & \multicolumn{1}{l}{} & \multicolumn{1}{l}{} & \multicolumn{1}{l}{} & \multicolumn{1}{l}{} \\
Depth map              & 0.0852               & 0.0862               & 0.0880               & 0.0722               & 0.0726               & 0.0742               \\ \midrule
\textit{GAN}                    & \multicolumn{1}{l}{} & \multicolumn{1}{l}{} & \multicolumn{1}{l}{} & \multicolumn{1}{l}{} & \multicolumn{1}{l}{} & \multicolumn{1}{l}{} \\
Depth map              & 0.0867               & 0.0955               & 0.0930               & 0.0799               & 0.0808               & 0.0878              \\
Grayscale              & 0.2238               & 0.1967               & 0.2018               & 0.1721               & 0.1845               & 0.1841               \\ \bottomrule
\end{tabular}
\end{table}

% We train the two configurations for output dimensions of $32\times 32$, $64\times 64$ and $128\times 128$ and report obtained loss on the test set in Table~\ref{tab:results-gan}. For a qualitative assessment, selected samples of the test set are shown in Fig.~\ref{fig:image}. 
% For grayscale image reconstruction we report obtained loss on the test set in Table~\ref{tab:results-gan} and show selected samples also in Fig.~\ref{fig:image}. %Both are shown for waveforms only

Table~\ref{tab:results-gan} compares the test set loss over a few design choices.  As in the ''Generator Only'' case, the loss is moderately higher for a larger depth map.  However,  Fig.~\ref{fig:image} shows our sample reconstructions by GAN have much finer details and clearer borders, and our grayscale reconstructions in the rightmost column have well placed floors even though objects are roughly outlined and abstracted. 
% {\setlength{\belowcaptionskip}{-3ex}
% \begin{figure}
%     \centering
%     %\vspace{3mm}
%     \includegraphics[width=\linewidth]{graphics/128_waveform.pdf}
%     \caption{\textit{Test samples shown for $128\times 128$ output resolution.} First and fourth column shows the ground truth depth map and grayscale image of the scene. The remaining columns show results from using raw waveforms as input. Generated depth-images show correct mapping of close and distant areas even in row three, where errors in the ground truth are present.}
%     \label{fig:image}
% \end{figure}
% }

% \begin{figure*}
%     \centering
%     \includegraphics[width=0.9\textwidth]{graphics/image.pdf}
%     \caption{``GAN'' grayscale test samples. First column shows the ground truth grayscale image of the scene. The remaining columns show results from using raw waveforms and spectrograms at resolutions $32\times 32$, $64\times 64$ and $128\times 128$.}
%     \label{fig:image}
% \end{figure*}

\subsection{Limitations of Our Approach}
How sound resonates, propagates and reflects in a room has a huge impact on sound-to-vision predictions.  
\begin{itemize}
\item Some materials have dampening properties, leading to faint or absorbed echos. 
\item Facing corners, where hallways fork in different directions, poses a big challenge, because sound waves scatter off in different directions.
\item At short ranges (e.g. \textless \SI{1}{\meter}), multi-path echoes could be received at the same time with similar amplitudes, creating a superposition that is difficult to resolve.
\item In areas with dense obstacles such as conference rooms with many office chairs, our sound-to-vision model often fails to predict any meaningful content (Fig.~\ref{fig:bad}).
\end{itemize}
%In addition, we fitted all sensors on a mobile robot to collect data from a perspective which will enable driving in the future but did not use the robot's own motor yet to minimize audible noise.

% This used to be called summary but it's really a conclusion
\section{Conclusions}
\label{sec:discussion}
Our BatVision system with a trained sound-to-vision model can reconstruct depth maps from binaural sound recorded by only two microphones to a remarkable accuracy.  It can predict detailed indoor scene depth and obstacles such as walls and furniture.  Sometimes, it  even outperforms our ground-truth depth map obtained from a stereo vision algorithm which struggles to estimate disparity  reliably.  

%In Fig.~\ref{fig:image} Row 3, per the actual room layout shown in the ground-truth grayscale scene image, our GAN result has the best depth map prediction, where corridors and open spaces can be distinguished and obstacles are visible, even though fine details are difficult to capture.
%in general the models perform well in its task to generative construct disparity like images where most detail of the scene is abstracted away. In some particular cases the predictions outperform the ground truth %as cameras are highly affected by visual conditions whereas sound waves are not. 
%due to the depth-from-stereo algorithm struggling to estimate the depth as a consequence of insufficient structures in both camera images.
%As can be seen in several examples in Fig.~\ref{fig:image}, the models perform particularly well in predicting ``open areas'' as to which areas of the scene are not occluded but is less precise when generating semantic detail of the actual scene.  

Generating the grayscale scene image is more difficult;  the amount of detail and information required is not expected to be present in sound echos.  However, our trained model is able to generate plausible wall placements and free floor areas.  When objects are not recognizable from the sound, the network fills in with an approximation of obstacles.  

Such seemingly incredible sound-to-vision results reflect natural statistical correlations between the sound and the image of indoor scenes, captured by our model trained on diverse scenes and likely utilized in a similar fashion by humans and animals.
%Note this is without being trained with any depth-related ground truth, \textit{i.e.} with monocular grayscale images only. 

%Finally, we do find classes of situations where the performance drops. Examples of this are shown in Fig.~\ref{fig:bad} and explained in the following:

% \begin{figure*}
%     \centering
%     \includegraphics[width=\textwidth]{graphics/bad.pdf}
%     \caption{``GAN'' test samples with poor result. First and second column show the grayscale image of the scene and ground truth depth map computed by the stereo camera. The remaining columns show results from using raw waveforms and spectrograms at resolutions $32\times 32$, $64\times 64$ and $128\times 128$.}
%     \label{fig:bad}
% \end{figure*}

% Because of the omnidirectional microphones, the method has the potential of producing \SI{360}{\degree} images. In this study our ground truth data was limited to the field of view of the camera, making the model suppress information about other directions. This could be improved in future work using a \SI{360}{\degree} camera.

\bibliographystyle{bib/IEEEtran}
\bibliography{bib/IEEEabrv,bib/bibliography}

\end{document}